\documentclass[letterpaper, 10 pt, conference]{ieeeconf}  

\IEEEoverridecommandlockouts                              

\usepackage{float}
\usepackage{graphics} 
\usepackage{epsfig} 
\usepackage{mathptmx} 
\usepackage{times} 
\usepackage{amsmath} 
\usepackage{amssymb}  
\usepackage{colortbl}
\usepackage[T1]{fontenc}
\usepackage{wrapfig,lipsum}
\usepackage{booktabs}
\usepackage{bigstrut}
\usepackage{multirow}
\usepackage{placeins}
\usepackage{gensymb}
\usepackage{xcolor}
\usepackage{soul}
\usepackage[ruled,linesnumbered, noend]{algorithm2e}
\DeclareMathOperator*{\argminB}{argmin}   
\usepackage{tikz}
\usepackage{yfonts}
\usepackage{graphicx}
\usepackage{import}
\usepackage{newtxtext}
\usepackage{newtxmath}
\usepackage{hhline}
\usepackage{stfloats}
\usepackage{atbegshi,picture} 
\usepackage{nccmath}
\usepackage{cite}
\usepackage[font={scriptsize}]{caption}
\makeatletter
\let\NAT@parse\undefined
\makeatother
\usepackage[hidelinks]{hyperref}  

\DeclareMathAlphabet\urwscr{U}{urwchancal}{m}{n}%
\DeclareMathAlphabet\rsfscr{U}{rsfso}{m}{n}
\DeclareMathAlphabet\euscr{U}{eus}{m}{n}
\DeclareFontEncoding{LS2}{}{}
\DeclareFontSubstitution{LS2}{stix}{m}{n}
\DeclareMathAlphabet\stixcal{LS2}{stixcal}{m} {n}

\newenvironment{myitem}{\begin{list}{$\bullet$}
{\setlength{\itemsep}{-0pt}
\setlength{\topsep}{0pt}
\setlength{\labelwidth}{5pt}
\setlength{\leftmargin}{10pt}
\setlength{\parsep}{-0pt}
\setlength{\itemsep}{0pt}
\setlength{\partopsep}{0pt}}}%
{\end{list}}

\title{\LARGE \bf BundleTrack: 6D Pose Tracking for Novel Objects\\ without Instance or Category-Level 3D Models
}

\author{Bowen Wen and Kostas Bekris 
\thanks{The authors are with the Computer Science Dept. of Rutgers in NJ, USA. Email: \{bw344,kostas.bekris\}@cs.rutgers.edu. This work is supported by NSF NRI award 1734492. The results do not express the sponsor's positions.}%
}

\begin{document}

\AtBeginShipoutNext{\AtBeginShipoutUpperLeft{%
  \put(\dimexpr\paperwidth-0.4cm\relax,-0.6cm){\makebox[0pt][r]{\framebox{IEEE/RSJ International Conference on Intelligent Robots and Systems (IROS) 2021}}}%
}}

\maketitle
\thispagestyle{empty}
\pagestyle{empty}

\begin{abstract}
Tracking the 6D pose of objects in video sequences is important for  robot manipulation. Most prior efforts, however, often assume that the target object's CAD model, at least at a category-level, is available for offline training or during online template matching. This work proposes \textit{BundleTrack}, a general framework for 6D pose tracking of novel objects, which does not depend upon 3D models, either at the instance or category-level. It leverages the complementary attributes of recent advances in deep learning for segmentation and robust feature extraction, as well as memory-augmented pose graph optimization for spatiotemporal consistency. This enables long-term, low-drift tracking under various challenging scenarios, including significant occlusions and object motions.  Comprehensive experiments given two public benchmarks demonstrate that the proposed approach significantly outperforms state-of-art, category-level 6D tracking or dynamic SLAM methods. When compared against state-of-art methods that rely on an object instance CAD model, comparable performance is achieved, despite the proposed method's reduced information requirements. An efficient implementation in CUDA provides a real-time performance of 10Hz for the entire framework. Code is available at: \url{https://github.com/wenbowen123/BundleTrack}

\end{abstract}

\section{INTRODUCTION}
\begin{figure}[t]
  \centering
  \includegraphics[width=0.48\textwidth]{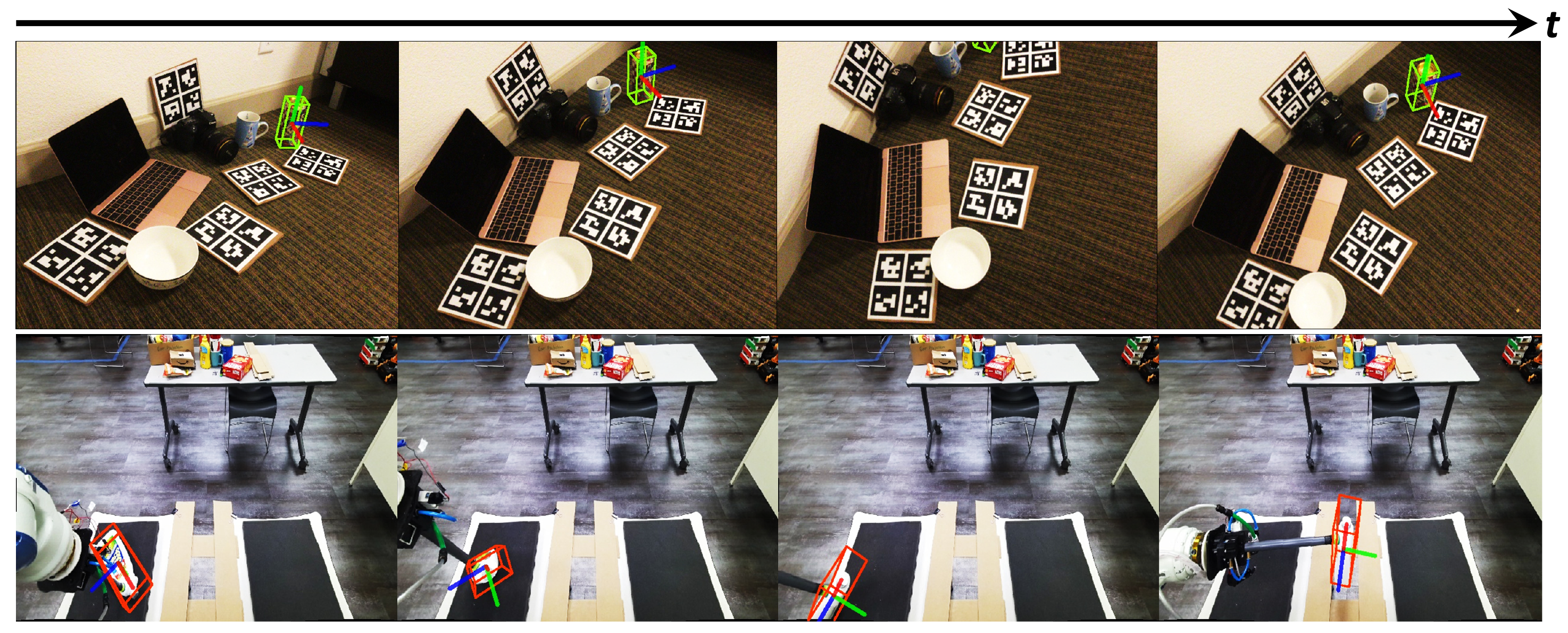}
  \vspace{-0.25in}
  \caption{\textbf{Top:} \textit{NOCS Dataset} \cite{Wang_2019_CVPR} example: The target object exits the camera's frustum during tracking but \textit{BundleTrack} maintains its estimate without re-initialization. \textbf{Bottom:} \textit{YCBInEOAT Dataset} \cite{wense3tracknet} example: The object is successfully tracked during pick and place manipulation by a robotic arm, despite the lack of texture, severe self-occlusion and motions due to the arm and the compliant suction cup. Computing object pose from forward kinematics is unreliable in this setup due to the end-effector.}
  \label{fig:intro}
  \vspace{-0.3in}
\end{figure}

Robot manipulation often requires information about the pose of the manipulated object. In some cases, this can be achieved through forward kinematics (FK), assuming the object's motion equivalent to the end-effector's motion. Frequently, however, FK is insufficient to accurately estimate the object's pose \cite{kappler2018real}. This can be due to slippage during grasping or in-hand manipulation \cite{wen2020robust}, or during handoffs or due to the compliance of a suction cup (Fig. \ref{fig:intro}). In these cases, dynamically estimating an object's pose from visual data is desirable. Single-image 6D pose estimation methods have been studied extensively \cite{xiang2017posecnn, park2019pix2pose, li2019cdpn, he2020pvn3d,mitash2020scene}. Some of them are fast and can re-estimate poses from scratch for every new frame \cite{tremblay2018deep, wang2019densefusion}. Nevertheless, this is redundant, less efficient, leading to less coherent estimations over consecutive frames and negatively impacts planning and control. On the other hand, given an initial pose estimate, tracking 6D object poses over image sequences can improve estimation speed while providing coherent and accurate poses by leveraging temporal consistency \cite{deng2019poserbpf,Wthrich2013ProbabilisticOT, schmidt2014dart}.

Most existing 6D object pose estimation or tracking approaches assume access to an  object instance's 3D model \cite{xiang2017posecnn,wang2019densefusion}. Having access to  such \textit{instance 3D models} complicates generalization to novel, unseen instances. To overcome this limitation, recent efforts have relaxed this assumption and require only \textit{category-level 3D models} for 6D pose estimation \cite{Wang_2019_CVPR,park2020latentfusion,chen2020category,chen2020learning} or tracking \cite{wang20196-pack}. They often achieve this by training over a large number of CAD models from the same category. While promising results have been demonstrated for previously seen object categories, there are still limitations. These methods are constrained by the variety of categories in the training database. Popular 3D model databases, such as \textit{ShapeNet} \cite{chang2015shapenet} and \textit{ModelNet40} \cite{wu20153d}, contain 55 and 40 categories respectively. This is still far from sufficient to cover diverse object categories present in the real world. Furthermore, 3D model databases often require nontrivial manual effort and expert domain knowledge to build, involving steps such as scanning \cite{newcombe2011kinectfusion}, mesh refinement \cite{cignoni2008meshlab} or CAD design.

Another line of work from the SLAM literature has moved to address dynamic, object-aware challenges \cite{xu2019mid, runz2018maskfusion, ma2015simultaneous, runz2017co}, where dynamic objects are being reconstructed on-the-fly while being tracked without the need for object 3D models beforehand. However, tracking-via-reconstruction \cite{runz2017co,runz2018maskfusion} tends to accumulate errors when fusing observations with erroneous pose estimates into the global model. These errors adversely impact model tracking in subsequent frames.

Motivated by the above limitations, this work aims for accurate, robust 6D pose tracking that is generalizable to novel objects without \textit{instance or category-level 3D models}. It exploits recent advances in video segmentation as well as learning-based keypoint detection and matching for a coarse pose estimate, followed by a memory-augmented pose-graph optimization step to achieve spatiotemporal consistent pose output. Instead of aggregating into a global model, representative historical observations are maintained as keyframes in a memory pool, providing candidate nodes for future graphs so as to enable multi-pair data association together with the latest observation. An efficient implementation of this framework in CUDA allows to achieve competitive running times. Extensive experiments have been conducted on two large-scale public benchmarks, shown in Fig. \ref{fig:intro}. Both qualitative and quantitative results demonstrate a significant improvement over existing state-of-art approaches, including methods using \textit{instance or category-level 3D models} or SLAM-like methods. 

In summary, this work's contributions are the following:


1) A novel integration of methods that result in a 6D pose tracking framework that generalizes to novel objects without access to instance or category-level 3D models.

2) A memory-augmented pose graph optimization for low-drift accurate 6D object pose tracking. In particular, augmenting the memory pool with historical observations enables multi-hop data association and ameliorate the dearth of correspondences between a pair of consecutive frames. Additionally, maintaining keyframes as raw nodes instead of aggregating into a global model significantly reduces tracking drift.

3) An efficient CUDA implementation, which allows to execute online the computationally-heavy multi-pair feature matching as well as pose-graph optimization for 6D object pose tracking (for the first time to the best of the authors' knowledge). 

These contributions result in a new state-of-art performance by boosting the previous best accuracy from \textbf{33.3\%} to \textbf{87.4\%} under the ``5\textdegree5cm'' metric in the \textit{NOCS Dataset} \cite{Wang_2019_CVPR}, even when compared against approaches utilizing category-level 3D models for training.  They also result in comparable performance on the \textit{YCBInEOAT} dataset \cite{wense3tracknet}, even when compared against approaches utilizing instance-level 3D models \cite{wense3tracknet}.

\section{RELATED WORK}
\textbf{6D Object Pose Tracking} - For setups where object CAD 
models are available, significant progress has been made in 6D pose tracking. This includes techniques based on hand-crafted probabilistic filtering \cite{choi2013rgb,Wthrich2013ProbabilisticOT,issac2016depth}, optimization \cite{schmidt2014dart, joseph2015versatile, zhong2019robust, tjaden2018region}, and machine learning \cite{deng2019poserbpf,wense3tracknet}. The requirements, however, of such \textit{instance-level 3D models}, either for training offline or model-frame registration during tracking, complicate generalization to novel instances. More recently,  a 6D pose tracking approach \cite{wang20196-pack} relaxed the assumption to \textit{category-level 3D models} using 3D object CAD model databases for training \cite{chang2015shapenet}. During testing, the target object category needs to be identified and the corresponding network for that category is utilized for tracking. Instead of being limited to the number of categories such database is able to include, this work employs deep features that in principle can be trained on arbitrary 2D images. It allows generalization to diverse novel objects, as shown in the accompanying experiments. 

\textbf{Dynamic Object-aware SLAM} - In order to track dynamic objects' pose and decouple them from static background, frame-model Iterative Closest Point \textit{(ICP)} combined with color \cite{xu2019mid,runz2018maskfusion,ma2015simultaneous,runz2017co}, probabilistic data association \cite{strecke2019fusion}, or 3D level-set likelihood maximization \cite{yuheng2013star3d} has been applied. Object models are simultaneously reconstructed on-the-fly by aggregating the observed RGB-D data with the newly tracked pose. Nevertheless, frame-model tracking can be challenging for object reconstruction, since errors in pose estimation transfer to the reconstructed model and adversely affect the subsequent tracking \cite{slavcheva2016sdf}. This work does not fuse observed frames but instead maintains them as nodes in a pose graph, allowing to correct previously erroneous estimates, and reduces drift in long-term tracking. The aforementioned SLAM-family approaches may also face challenges in robot manipulation setups that involve small, textureless, flat or shiny objects due to the dearth of sufficient correspondences between the pair of consecutive frames. To ameliorate this issue, \textit{BundleTrack} searches correspondences among current and multiple historical frames, consisting of both feature and geometric terms, as the edges in the pose graph. Its effectiveness has been shown in extensive experiments including for such challenging manipulation scenarios.

\textbf{3D Hand-held Object Scanning} - Promising results have been demonstrated in scanning dynamic hand-held objects \cite{tzionas20153d,weise2011online,krainin2010manipulator,wang2019hand,weise2008accurate}, where the object's motion needs to be taken into account similar to the current setup. In particular, a framework for robot manipulation \cite{krainin2010manipulator} performs simultaneous object reconstruction and tracking, which leads to similar issues as the aforementioned dynamic SLAM methods. In addition, forward kinematics is required in its Kalman Filtering framework, preventing generalization in scenarios when objects are not held by the robotic manipulator. While estimating object poses is part of the scanning process, there are key differences from online 6D pose tracking. For the scanning application, external assistance including human interaction or deliberate motion is acceptable \cite{weise2011online,wang2019hand,weise2008accurate} but it is not assumed in the current work. Furthermore, time consuming global-optimization steps are often adopted at the end of scanning to polish the models and their poses while intermediate erroneous pose estimations and associated frames can be discarded and not fused into the global model \cite{weise2011online,wang2019hand,weise2008accurate}. In contrast, this work aims to provide fast and accurate pose tracking output online.

\section{PROBLEM FORMULATION}
\begin{figure*}[h!]
  \centering
  \vspace{+0.05in}
  \includegraphics[width=\textwidth]{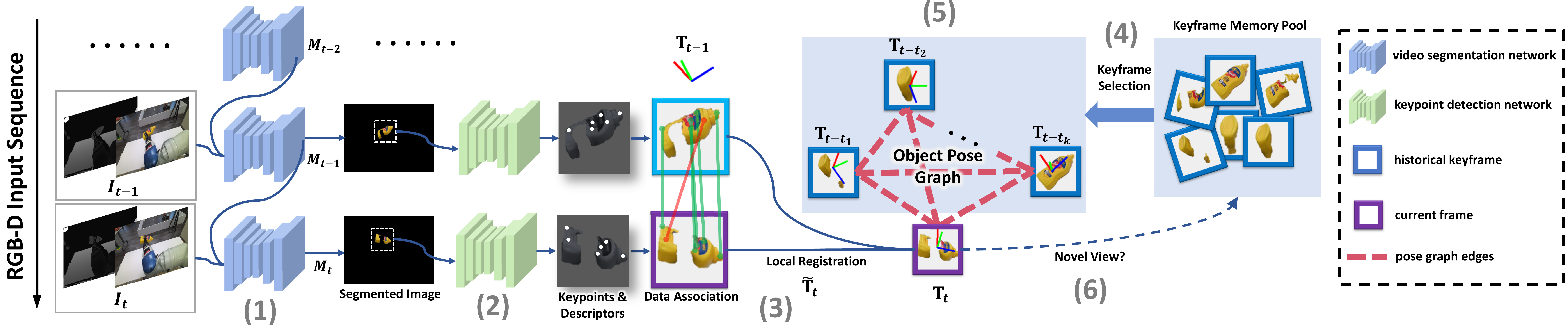}
  \vspace{-0.25in}\caption{\textit{BundleTrack} framework from left to right: (1) an image segmentation network returns the object mask given the prior one; (2) a network detects keypoints and their descriptors; (3) keypoints are matched and coarse registration is performed between consecutive frames to estimate an initial relative transform $\tilde{\mathbf{T}}_t$; (4) keyframes are selected from a memory pool to participate in the pose graph optimization; (5) online pose graph optimization outputs a refined spatiotemporal consistent pose $\mathbf{T}_t$; and (6) the latest frame is included in the memory pool, if it is a novel view to enrich diversity.}
  \label{fig:pipeline}
  \vspace{-0.2in}
\end{figure*}

\setlength{\columnsep}{0.1in}
\setlength{\intextsep}{0.03in}
\begin{wrapfigure}{r}{1.9in}
\vspace{-.1in}
  \centering
  \includegraphics[width=1.9in]{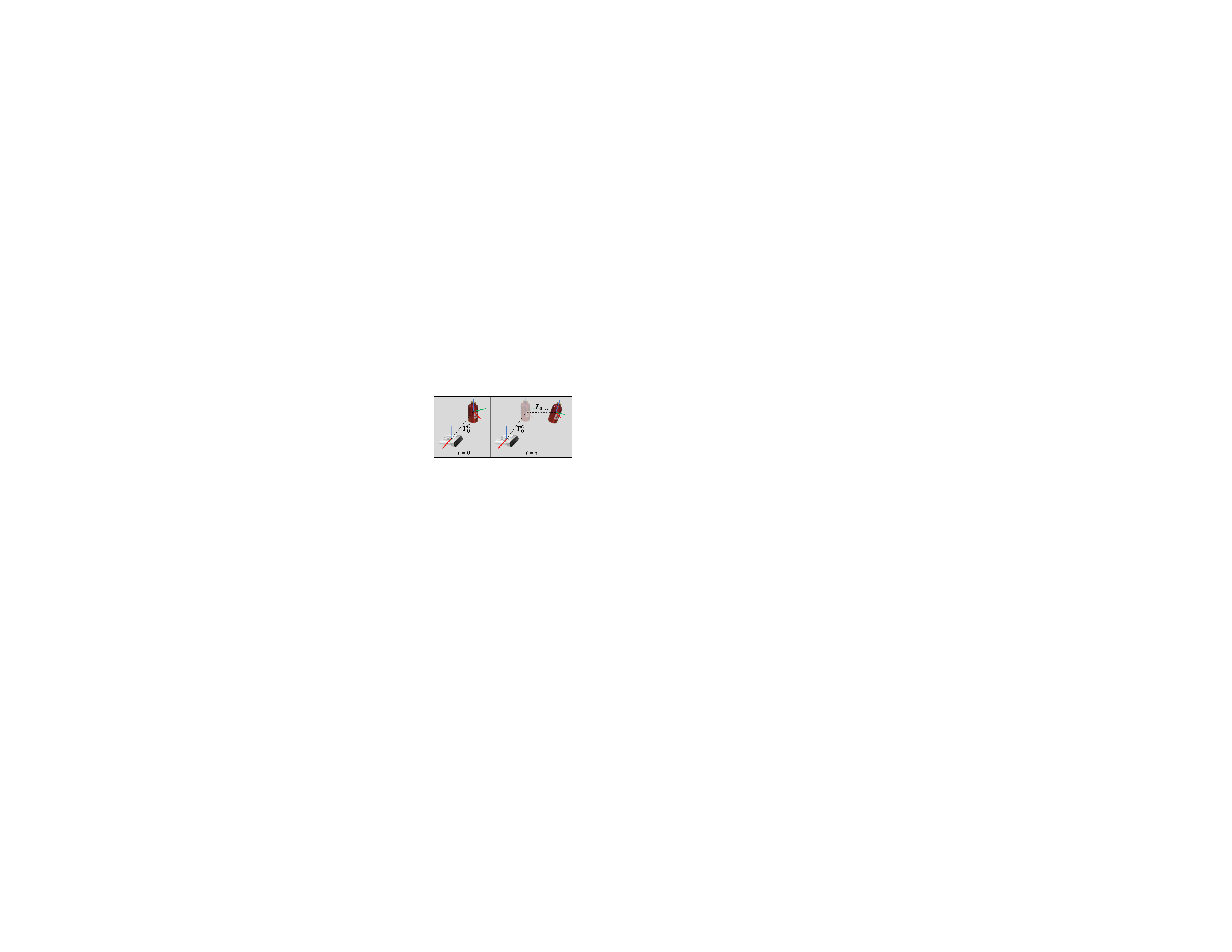}  
\end{wrapfigure}

Assume a rigid body for which there is no its corresponding 3D model, nor its category-level 3D model database for training. The objective is to continuously track its 6D pose change relative to the start of tracking, i.e., the relative transformation $ T_{0 \rightarrow \tau} \in SE(3), \tau \in \{1,2,...,t\}$ in the camera's frame $C$. The input is the following:

\begin{myitem}
  \item $I_{\tau}$: A sequence of RGB-D data $I_{\tau}, \tau \in \{0,...,t\}$.
  \item $M_0$: A binary mask on the first image $I_0$, indicating the target object region to track in the image space. 
  \item $T_{0}^{C}$ (optional):  The initial pose in the camera's frame $C$. Used if the objective is to recover the object's absolute pose in $C$, otherwise set to identity.
\end{myitem}
    
\noindent The initial mask $M_0$ can be obtained in multiple different ways to initialize tracking. For instance, via semantic segmentation  \cite{long2015fully,chen2017deeplab,le2018deep} or non-semantic methods, such as image segmentation, \cite{meyer1992color,danielczuk2019segmenting,xiang2020learning}, point cloud segmentation/clustering \cite{rusu20113d,Papon13CVPR}, or plane fitting and removal \cite{rusu20113d}, etc.
    

The object's pose in the camera's frame $C$ can be recovered at any timestamp by applying the relative transformation $T_{0 \rightarrow \tau}$ in the camera's frame $\mathbf{T}_{\tau} = T^{C}_{\tau} = T_{0}^{C}[(T_{0}^{C})^{-1} T_{0 \rightarrow \tau} T_{0}^{C}] = T_{0 \rightarrow \tau} T_{0}^{C} \in SE(3)$. For simplicity, the rest of this document will refer to $\mathbf{T}_{\tau}$ as the output of the process but $T_{0 \rightarrow \tau}$ is what is actually computed as tracking.

\section{APPROACH}


An overview of the proposed \textit{BundleTrack} framework is depicted in Fig. \ref{fig:pipeline}. The currently observed RGB-D frame $I_{t}$ and the object segmentation mask computed during the last timestamp $M_{t-1}$  are forwarded to a video segmentation network to compute the current object mask $M_{t}$. Based on $M_t$ and $M_{t-1}$ respectively, the target object regions in both $I_{t}$ and $I_{t-1}$ are cropped, resized and sent to a keypoint detection network to compute keypoints and feature descriptors. A data association process consisting of feature matching and outlier pruning in the manner of \textit{RANSAC} \cite{fischler1981random} identifies feature correspondences. Based on these correspondences, a registration between $I_{t-1}$ and $I_{t}$ can be solved in closed-form, which is then used to provide a coarse estimate $\tilde{\mathbf{T}}_t$ for the transform between the two snapshots. The estimate $\tilde{\mathbf{T}}_t$ is used to initialize the current node $\mathbf{T}_{t}$ as part of a pose graph optimization step. To define the rest of the nodes of the pose graph, no more than $\mathcal{K}$ keyframes are selected from a memory pool to participate in the optimization. The choice of $\mathcal{K}$ is made to balance an efficiency vs. accuracy tradeoff. Pose graph edges include both feature and geometric correspondences, which are computed in parallel on GPU. Given this information, the pose graph step outputs online the optimized pose for the current timestamp $\mathbf{T}_{t} \in SE(3)$. If the last frame corresponds to a novel view, then it is also included in the memory pool.

\subsection{Propagating Object Segmentation}
\label{sec:video_seg}


The first step is to segment the object's image region from the background. Prior work \cite{runz2018maskfusion} used \textit{Mask-RCNN} \cite{he2017mask} to compute the object mask in every frame of the video. It deals with each new frame independently, which is less efficient and results in temporal inconsistencies. 

To avoid these limitations, this work adopts an off-the-shelf \textit{transductive-VOS} network \cite{zhang2020a} for video object segmentation, which is trained on the \textit{Davis 2017} \cite{pont20172017} and \textit{Youtube-VOS} \cite{xu2018youtube} datasets.  The network uses dense long-term similarity dependencies between current and past feature embeddings to propagate the previous object mask to the latest frame. The object mask needed by \textit{BundleTrack} is simply binary, i.e., $M_{\tau}=\{0,1\}^{H \times W}, \tau \in \{0,1,...,t\}$ and distinguishes the object region from the background.  The only requirement is an initial mask $M_{0}$ of interest. Neither the \textit{transductive-VOS} network nor the following steps of \textit{BundleTrack} require $M_{0}$ to come from semantic/instance segmentation. Therefore, it can also be obtained in  alternative ways depending on the application, e.g., low-level image segmentation \cite{meyer1992color,lafferty2001conditional}, point cloud segmentation/clustering \cite{rusu20113d,Papon13CVPR}, or plane fitting and removal \cite{rusu20113d}, etc.  



While the current implementation uses \textit{transductive-VOS}, the following techniques do not depend on this specific network. If the object mask can be computed via simpler means, such as computing a region of interest (ROI) from forward kinematics followed by point cloud filtering in robot manipulation scenarios \cite{wen2020robust}, the segmentation module can be replaced.

\vspace{-0.05in}
\subsection{Keypoint Detection, Matching and Local Registration} 
\label{sec:local_registration}
\vspace{-0.05in}

Local registration is performed between consecutive frames $I_{t-1}$ and $I_t$ to compute a initial pose $\tilde{\mathbf{T}}_t$. To do so, correspondence between keyframes detected on each image is performed. Different from prior work \cite{wang20196-pack}, which relies on \textit{category-level 3D models} to learn a fixed number of category-level semantic keypoints, this work aims to use generalizable features not specific to certain instances or categories. The \textit{LF-Net} \cite{ono2018lf} is chosen given its satisfactory balance between performance and inference speed. It only requires training on general 2D images, such as the \textit{ScanNet dataset} \cite{dai2017scannet} used here, and generalizes to novel scenes. During testing, for the newly observed frame $I_t$, \textit{LF-Net} receives the segmented image (Sec. \ref{sec:video_seg}) as input. It then outputs $n$ keypoints $x_i, i \in \{0,1,...,n-1\}$ along with the feature descriptor $D_i \in R^{128}$, where $n$ is 500 in all experiments. Due to the potentially imperfect segmentation in previous step, outlier keypoints can arise from the background. It is thus critical to perform feature matching and outlier pruning via \textit{RANSAC} \cite{fischler1981random}, executed in parallel on GPU in this work. Each registration sample consists of 3 pairs of keypoints matched between the two images. A pose hypothesis is generated from a sample via least squares \cite{arun1987least}. When evaluating samples, inlier correspondences have a distance between transformed point pairs below a threshold $\delta$ and an angle formed by the normals within a threshold $\alpha$. The values of $\delta$ and $\alpha$ are empirically set to $5mm$ and $45\degree$ in all experiments. After \textit{RANSAC}, a preliminary pose is computed by $\tilde{\mathbf{T}}_t=\mathbf{T}_{t-1} T_{t}^{t-1}$ where $T_{t}^{t-1}$ is the best sampled correspondence hypothesis. 


\subsection{Keyframe Selection}
\label{sec:keyframe_selection}

$\tilde{\mathbf{T}}_t$ is then refined during a pose graph optimization step. The number of keyframes participating in the  optimization is limited to $k\leqslant \mathcal{K}$ for the sake of efficiency, where $\mathcal{K}=15$ is the number used in the experiments. When the size of the keyframe memory pool $\mathcal{N}$ is larger than $\mathcal{K}$, the objective is to find the set of keyframes with the largest mutual viewing overlap to make good use of multi-view consistency. This challenge can be formulated as the minimum H-subgraph of an edge-weighted graph problem \cite{vassilevska2006finding}:

\vspace{-0.02in}
\begin{gather*}
\argminB_{x} \ \sum\limits_{i\in \mathcal{N}}\sum\limits_{j\in \mathcal{N}, j\neq i} x_{i}x_{j} \cdot arccos \left( \frac{tr(R_{i}^{T}R_{j})-1}{2} \right)\\
\textrm{so that}: \sum\limits_{i\in \mathcal{N}} x_{i}=\mathcal{K}  \textrm{ and } x_{i}\in \{0,1\}, i\in \mathcal{N},
\end{gather*}

\noindent where $R_i$ is the rotation matrix of the corresponding keyframe's pose. The goal is to find the optimal binary vector $x\in R^{N}$ that indicates the selections. The weight of the edge between frame pair $(i,j)$ is the geodesic distance of their rotations. Mutual viewing overlap is maximized when the mutual rotation difference relative to the camera is minimized. Combinatorial optimization algorithms for solving this problem have a complexity of $O(\mathcal{N}^{\mathcal{K}}/ log \mathcal{N})$ \cite{vassilevska2006finding}. In practice, an iterative greedy selection is followed by starting with the keyframe set $\{I_{0}\}$ until the number of selected keyframes reaches $\mathcal{K}$. $I_{0}$ is chosen since the initial frame does not suffer from any tracking drift and serves as the reference frame. In each iteration, the keyframe with the smallest sum of geodesic distances against $I_t$ as well as all previously selected keyframes is added. This reduces complexity to $O(\mathcal{N} \mathcal{K}^{3}+\mathcal{N} \mathcal{K}^{2})$, making the selection practical (under a millisecond) without degrading performance.

\subsection{Online Pose Graph Optimization}
\label{sec:pose_graph}

The pose graph can be denoted as $G=\{V,E\}, |V|=k+1$, where each node corresponds to the object pose in the camera's frame at the current and $k$ selected timestamps $\tau \in \{t,t-t_{1},t-t_{2},...,t-t_{k}\}$. For simplicity, the subscripts of graph nodes will be denoted as simple indices $i \in |V|$ instead of the actual timestamp $t-t_{i}$.  Each node's pose can then be denoted as $\mathbf{T}_{i}, i \in |V|$.  Inspired by \cite{dai2017bundlefusion}, for the edges between each pair of nodes, two types of energies $\mathbf{E}_{f}$ and $\mathbf{E}_{g}$ are considered. The energy $\mathbf{E}_{f}$ relates to the residuals computed from feature correspondences and $\mathbf{E}_{g}$ relates to the geometric residuals measured by dense pixel-wise point-to-plane distance. The spatiotemporal consistency is achieved when the total energy of the graph $\mathbf{E}$ is minimized:
\begin{equation}\label{eq:E}
\vspace{-0.in}
\scalebox{1.0}{
$
\mathbf{E}= \sum\limits_{i \in |V|} \sum\limits_{j \in |V|, j\neq i} (\lambda_{1} \mathbf{E}_{f}(i,j) + \lambda_{2} \mathbf{E}_{g}(i,j))
$
}
\end{equation}
\begin{equation}\label{eq:E_feat}
\scalebox{1.0}{
$
\mathbf{E}_{f}(i,j)=\sum\limits_{(m,n) \in C_{i,j}} \rho \left( \left \| \mathbf{T}_{i}^{-1}p_{m} - \mathbf{T}_{j}^{-1}p_{n} \right \|_{2} \right)
$
}
\end{equation}

In order to compute $\mathbf{E}_{f}$, feature correspondences $C_{i,j}$ between each pair of nodes $(i,j)$ are determined. If $C_{i,j}$ has been built during a previous pose graph optimization, it is reused. Otherwise, the data association process of Sec. \ref{sec:local_registration} is performed to compute $C_{i,j}$. These multi-pair feature correspondences are built in parallel on GPU. In Eq. (\ref{eq:E_feat}) and (\ref{eq:E_geom}), $p$ represents the unprojected 3D points in the camera's frame, $\rho$ is the M-estimator, where Huber loss is used. 

\begin{equation}\label{eq:E_geom}
\scalebox{1.0}{
$
\hspace{-0.12in} \mathbf{E}_{g}(i,j)=  \sum\limits_{p\in |I_i|}\rho \left( \left\|  n_i(x) \cdot ( \mathbf{T}_{i}  \mathbf{T}_{j}^{-1}\pi^{-1}_{D} (\pi(\mathbf{T}_{j} \mathbf{T}_{i}^{-1}p)) - p)    \right\|_{2} \right)
$
}
\end{equation}

For $\mathbf{E}_{g}$, dense pixel-wise correspondences are associated by point re-projection, while outliers are filtered based on the distance between the point pair and the angle formed by their normals; $\pi(\cdot)$ is the perspective projection operation; $\pi^{-1}_{D}(\cdot)$ denotes the unprojection mapping, which recovers a 3D point in the camera's frame by looking up the depth value on the pixel location; $n_{i}(\cdot)$ returns the normal of the pixel on the frame $I_{i}, i \in |V|$. 

In Eq. (\ref{eq:E}),  $\lambda_{1}$ and $\lambda_{2}$ are the weights balancing $\mathbf{E}_{f}$ and $\mathbf{E}_{g}$. To emphasize the lack of sensitivity to the choice of these values, $\lambda_{1}$ and $\lambda_{2}$ are set to $1$ in all experiments unless otherwise specified. Then, the goal is to find the optimal poses, such that:
\vspace{-0.in}
$$
\xi^{*}=\argminB_{\xi}  \rho ( \mathbf{\bar{E}(\xi)} ) 
$$
where $\mathbf{\bar{E}(\xi)}$ is the stacked energy residual vector, $\mathbb{\xi}=(\xi_t, \xi_{t-t_1}, \xi_{t-t_2}, ..., \xi_{t-t_k})^{T} \in R^{6 \times (k+1)}$ is the stacked pose vector corresponding to the current frame and $k$ selected past keyframes, while the pose corresponding to the initial frame $I_0$ is kept constant as reference. Each block $\xi_i=log(\mathbf{T}_i) \in \mathfrak{se}(3)$ is parametrized in Lie Algebra \cite{bourbaki2008lie}, consisting of 3 parameters for translation and 3 parameters for rotation. A common approach is to apply first-order Taylor expansion around $\xi$, such that the iteratively re-weighted nonlinear least squares can be solved by a Gauss-Newton update:
$$
(\mathbf{J}^{T}\mathbf{W}\mathbf{J}) \Delta \xi = \mathbf{J}^T \mathbf{W} \mathbf{\bar{E}}
$$
where $\mathbf{J}$ is the Jacobian matrix with respect to $\xi$, $\mathbf{W}$ is a diagonal weight matrix computed by the M-estimator $\rho$ and residual, which is updated in each iteration. To better take advantage of the sparsity of $\mathbf{J}$ and $\mathbf{W}$, inside each Gauss-Newton step, an iterative PCG (Preconditioned Conjugate Gradient) \cite{hildebrand1987introduction} solver is leveraged, where the diagonal matrix $\mathbf{J}^{T}\mathbf{W}\mathbf{J}$ is used as the preconditioner. Incremental pose updates are accumulated in the tangent space after each iteration $\xi \leftarrow \xi \boxplus \Delta \xi$. The entire pose graph optimization is implemented in CUDA for parallel computation. 

At the end of the optimization, the object pose corresponding to each graph node is obtained by $\mathbf{T}_{i}=exp(\xi_i) \in SE(3), i \in |V|$. The one corresponding to the current timestamp $t$ becomes the output tracked pose $\mathbf{T}_{t}$, while poses corresponding to the historical keyframes are updated in the memory pool. The entire process is causal, i.e. past frames' corrected poses cannot be updated in the output. However, their corrected pose estimates provide better initialization in following pose graph optimization steps to benefit the solution of new observations. This significantly reduces long-term drift compared against tracking-via-reconstruction \cite{runz2018maskfusion}, where any intermediate erroneous pose estimation introduces noise when fused into the global model and adversely affects the subsequent tracking. 

\subsection{Augmenting the Keyframe Memory Pool}
The initial frame $I_0$ is always selected as it does not suffer from any tracking drift. For later frames, once the current object pose $\mathbf{T}_{t}$ is determined, its rotation geodesic distance against each existing keyframe in the pool is compared. If all pair-wise distances are larger than $\alpha$ ($arccos(10\degree)$ in all experiments), $I_{t}$ is added into the keyframe memory pool. This encourages to add frames from novel views, such that multi-view diversity is enriched. 

\section{EXPERIMENTS}
This  section  evaluates  the  proposed  approach  and  compares  against state-of-the-art 6D pose tracking and estimation methods on two public benchmarks, the \textit{NOCS dataset} \cite{Wang_2019_CVPR} and the \textit{YCBInEOAT dataset} \cite{wense3tracknet}. Experiments are performed over diverse types of objects and various tracking scenarios (e.g., moving camera or moving objects). Both quantitative and qualitative results demonstrate that \textit{BundleTrack} achieves comparable or even superior performance relative to alternatives, although it does not require \textit{instance or category-level 3D models}. Concretely, no CAD models or training data from a 3D object database are used by \textit{BundleTrack}. All  experiments are  conducted on a standard desktop with Intel  Xeon(R) E5-1660 v3@3.00GHz  processor and a single NVIDIA RTX 2080 Ti GPU. 

\subsection{Datasets}

\noindent \textbf{NOCS dataset \cite{Wang_2019_CVPR}:} Among existing datasets, this is the closest to the setup here, where \textit{instance 3D models} are not provided during evaluation. The dataset contains 6 object categories: bottle, bowl, camera, can, laptop, and mug.  The training set consists of: (1) 7 real videos containing 3 instances of each category in total, annotated with ground truth poses; and (2) 275K frames of synthetic data generated using 1085 instances from the above 6 categories using a 3D model database \textit{ShapeNetCore} \cite{chang2015shapenet} with random poses and object combinations in each scene.  The testing set has 6 real videos containing 3 different unseen instances within each category, resulting in 18 different object instances and 3,200 frames in total.

\noindent \textbf{YCBInEOAT dataset \cite{wense3tracknet}:} This dataset helps verify the effectiveness of 6D pose tracking during robot manipulation. It was originally developed to evaluate approaches relying on CAD models. The available CAD models, however, are not used by \textit{BundleTrack}. In contrast to the \textit{NOCS dataset} where objects are statically placed on a tabletop and captured by a moving camera, \textit{YCBInEOAT} contains 9 video sequences captured by a static RGB-D camera, while objects are dynamically manipulated. There are three types of manipulation: (1) single arm pick-and-place, (2) within-hand manipulation, and (3) pick to hand-off between arms to placement. These scenarios and the end-effectors used make directly computing poses from forward kinematics unreliable.  The manipulation videos involve 5 \textit{YCB Objects} \cite{calli2015benchmarking}: mustard bottle, tomato soup can, sugar box, bleach cleanser and cracker box. 

\subsection{Results on the NOCS Dataset} 
\label{sec:eval_nocs}

\begin{table}
\centering
\vspace{+0.1in}
\resizebox{0.48\textwidth}{!}{

\begin{tabular}{c|c|l|rrrrrr||r}
\hline
\textbf{Assumption} &
  \textbf{Methods} &
  \multicolumn{1}{c|}{\textbf{Metrics}} &
  bottle &
  bowl &
  camera &
  can &
  laptop &
  mug &
  Overall
  \bigstrut\\
\hline
\multicolumn{1}{c|}{\multirow{16}[8]{*}{\textbf{\shortstack{Category\\-Level\\3D Model}}}} &
  \multirow{4}[2]{*}{NOCS \cite{Wang_2019_CVPR}} &
  5\textdegree5cm &
  5.5 &
  \textbf{62.2} &
  0.6 &
  7.1 &
  25.5 &
  0.9 &
  17.0
  \bigstrut[t]\\
 &
   &
  IoU25 &
  48.7 &
  99.6 &
  90.6 &
  77.0 &
  94.7 &
  82.8 &
  82.2
  \\
 &
   &
  R\textsubscript{err} &
  25.6 &
  \textbf{4.7} &
  \textbf{33.8} &
  16.9 &
  8.6 &
  31.5 &
  20.2
  \\
 &
   &
  T\textsubscript{err} &
  14.4 &
  \textbf{1.2} &
  \textbf{3.1} &
  \textbf{4.0} &
  \textbf{2.4} &
  4.0 &
  4.9
  \bigstrut[b]\\
\cline{2-10} &
  \multirow{4}[2]{*}{\shortstack{KeypointNet\\\cite{suwajanakorn2018discovery}}} &
  5\textdegree5cm &
  5.9 &
  16.8 &
  1.8 &
  4.3 &
  49.2 &
  3.1 &
  13.5
  \bigstrut[t]\\
 &
   &
  IoU25 &
  23.1 &
  74.7 &
  30.9 &
  42.6 &
  94.6 &
  52.0 &
  53.0
  \\
 &
   &
  R\textsubscript{err} &
  28.5 &
  9.8 &
  45.2 &
  28.8 &
  6.5 &
  61.2 &
  30.0
  \\
 &
   &
  T\textsubscript{err} &
  9.5 &
  8.2 &
  8.5 &
  13.1 &
  4.4 &
  6.7 &
  8.4
  \bigstrut[b]\\
\cline{2-10} &
  \multicolumn{1}{c|}{\multirow{4}[2]{*}{\shortstack{6-PACK w/o\\\newline{}temporal \cite{wang20196-pack}}}} &
  5\textdegree5cm &
  23.7 &
  53.0 &
  8.4 &
  \textbf{25.0} &
  62.4 &
  22.4 &
  32.5
  \bigstrut[t]\\
 &
   &
  IoU25 &
  \textbf{92.0} &
  \textbf{100.0} &
  \textbf{91.0} &
  89.9 &
  97.8 &
  \textbf{100.0} &
  \textbf{95.1}
  \\
 &
   &
  R\textsubscript{err} &
  15.7 &
  5.3 &
  43.9 &
  \textbf{12.5} &
  4.9 &
  \textbf{20.3} &
  17.1
  \\
 &
   &
  T\textsubscript{err} &
  4.2 &
  1.6 &
  5.5 &
  5.0 &
  2.5 &
  \textbf{1.8} &
  \textbf{3.4}
  \bigstrut[b]\\
\cline{2-10} &
  \multirow{4}[2]{*}{6-PACK \cite{wang20196-pack}} &
  5\textdegree5cm &
  \textbf{24.5} &
  55.0 &
  \textbf{10.1} &
  22.6 &
  \textbf{63.5} &
  \textbf{24.1} &
  \textbf{33.3}
  \bigstrut[t]\\
 &
   &
  IoU25 &
  91.1 &
  \textbf{100.0} &
  87.6 &
  \textbf{92.6} &
  \textbf{98.1} &
  95.2 &
  94.2
  \\
 &
   &
  R\textsubscript{err} &
  \textbf{15.6} &
  5.2 &
  35.7 &
  13.9 &
  \textbf{4.7} &
  21.3 &
  \textbf{16.0}
  \\
 &
   &
  T\textsubscript{err} &
  \textbf{4.0} &
  1.7 &
  5.6 &
  4.8 &
  2.5 &
  2.3 &
  3.5
  \bigstrut[b]\\
\hline
\hline
\multirow{16}[8]{*}{\textbf{\shortstack{No\\Model}}} &
  \multirow{4}[2]{*}{ICP \cite{zhou2018open3d}} &
  5\textdegree5cm &
  10.1 &
  40.3 &
  12.6 &
  17.2 &
  14.8 &
  6.2 &
  16.9
  \bigstrut[t]\\
 &
   &
  IoU25 &
  29.9 &
  79.7 &
  53.1 &
  40.5 &
  50.9 &
  27.7 &
  47.0
  \\
 &
   &
  R\textsubscript{err} &
  48.0 &
  19.0 &
  80.5 &
  47.1 &
  37.7 &
  56.3 &
  48.1
  \\
 &
   &
  T\textsubscript{err} &
  15.7 &
  4.7 &
  12.2 &
  9.4 &
  9.2 &
  9.2 &
  10.5
  \bigstrut[b]\\
\cline{2-10} &
  \multirow{4}[2]{*}{\shortstack{TEASER++\textsuperscript{*} \\\cite{Yang20troteaser}}} &
  5\textdegree5cm &
  13.9 &
  35.5 &
  10.7 &
  11.7 &
  40.9 &
  7.5 &
  20.0
  \bigstrut[t]\\
 &
   &
  IoU25 &
  \textbf{100.0} &
  \textbf{99.9} &
  \textbf{99.9} &
  \textbf{100.0} &
  \textbf{99.9} &
  \textbf{99.9} &
  \textbf{99.9}
  \\
 &
   &
  R\textsubscript{err} &
  17.0 &
  10.6 &
  18.8 &
  20.4 &
  7.2 &
  23.0 &
  16.2
  \\
 &
   &
  T\textsubscript{err} &
  2.7 &
  1.8 &
  2.8 &
  2.7 &
  2.6 &
  2.4 &
  2.5
  \bigstrut[b]\\
\cline{2-10} &
  \multirow{4}[2]{*}{\shortstack{MaskFusion\\\cite{runz2018maskfusion}}} &
  5\textdegree5cm &
  15.5 &
  32.3 &
  11.7 &
  8.8 &
  73.9 &
  16.4 &
  26.5
  \bigstrut[t]\\
 &
   &
  IoU25 &
  51.4 &
  71.4 &
  60.8 &
  49.7 &
  \textbf{99.9} &
  56.2 &
  64.9
  \\
 &
   &
  R\textsubscript{err} &
  36.7 &
  12.3 &
  43.0 &
  34.9 &
  3.4 &
  40.6 &
  28.5
  \\
 &
   &
  T\textsubscript{err} &
  11.3 &
  5.3 &
  11.1 &
  9.3 &
  3.5 &
  9.2 &
  8.3
  \bigstrut[b]\\
\cline{2-10} &
  \multicolumn{1}{c|}{\multirow{4}[2]{*}{\shortstack{BundleTrack\\(Ours)}}} &
  5\textdegree5cm &
  \cellcolor[rgb]{ .851,  .851,  .851}\textbf{86.5} &
  \cellcolor[rgb]{ .851,  .851,  .851}\textbf{99.6} &
  \cellcolor[rgb]{ .851,  .851,  .851}\textbf{85.8} &
  \cellcolor[rgb]{ .851,  .851,  .851}\textbf{99.2} &
  \cellcolor[rgb]{ .851,  .851,  .851}\textbf{99.9} &
  \cellcolor[rgb]{ .851,  .851,  .851}\textbf{53.6} &
  \cellcolor[rgb]{ .851,  .851,  .851}\textbf{87.4}
  \bigstrut[t]\\
 &
   &
  IoU25 &
  \cellcolor[rgb]{ .851,  .851,  .851}\textbf{100.0} &
  \cellcolor[rgb]{ .851,  .851,  .851}\textbf{99.9} &
  \cellcolor[rgb]{ .851,  .851,  .851}\textbf{99.9} &
  \cellcolor[rgb]{ .851,  .851,  .851}\textbf{100.0} &
  \cellcolor[rgb]{ .851,  .851,  .851}\textbf{99.9} &
  \cellcolor[rgb]{ .851,  .851,  .851}\textbf{99.9} &
  \cellcolor[rgb]{ .851,  .851,  .851}\textbf{99.9}
  \\
 &
   &
  R\textsubscript{err} &
  \cellcolor[rgb]{ .851,  .851,  .851}\textbf{1.6} &
  \cellcolor[rgb]{ .851,  .851,  .851}\textbf{1.7} &
  \cellcolor[rgb]{ .851,  .851,  .851}\textbf{3.0} &
  \cellcolor[rgb]{ .851,  .851,  .851}\textbf{1.5} &
  \cellcolor[rgb]{ .851,  .851,  .851}\textbf{1.5} &
  \cellcolor[rgb]{ .851,  .851,  .851}\textbf{5.2} &
  \cellcolor[rgb]{ .851,  .851,  .851}\textbf{2.4}
  \\
 &
   &
  T\textsubscript{err} &
  \cellcolor[rgb]{ .851,  .851,  .851}\textbf{2.3} &
  \cellcolor[rgb]{ .851,  .851,  .851}\textbf{2.1} &
  \cellcolor[rgb]{ .851,  .851,  .851}\textbf{2.1} &
  \cellcolor[rgb]{ .851,  .851,  .851}\textbf{2.1} &
  \cellcolor[rgb]{ .851,  .851,  .851}\textbf{2.2} &
  \cellcolor[rgb]{ .851,  .851,  .851}\textbf{2.2} &
  \cellcolor[rgb]{ .851,  .851,  .851}\textbf{2.1}
  \bigstrut[b]\\
\hline
\end{tabular}%

}
\vspace{-0.05in}
\caption{Results on the \textit{NOCS dataset} \cite{Wang_2019_CVPR}. For the metrics of 5\textdegree5cm and IoU25, a higher value is preferable. For the metrics of R\textsubscript{err} and T\textsubscript{err}, a lower value is preferable. Under each type of 3D model assumption, the best results are highlighted in bold font. TEASER++\textsuperscript{*} denotes TEASER++ \cite{Yang20troteaser} operating over the same segmented point cloud and feature correspondences as in the proposed \textit{BundleTrack}.}
\label{tab:nocs}
\vspace{-0.2in}
\end{table}

\begin{figure}[h]
  \centering
  \definecolor{blue}{RGB}{0,112,192}
  \definecolor{green}{RGB}{0, 176, 80}
  \includegraphics[width=0.48\textwidth]{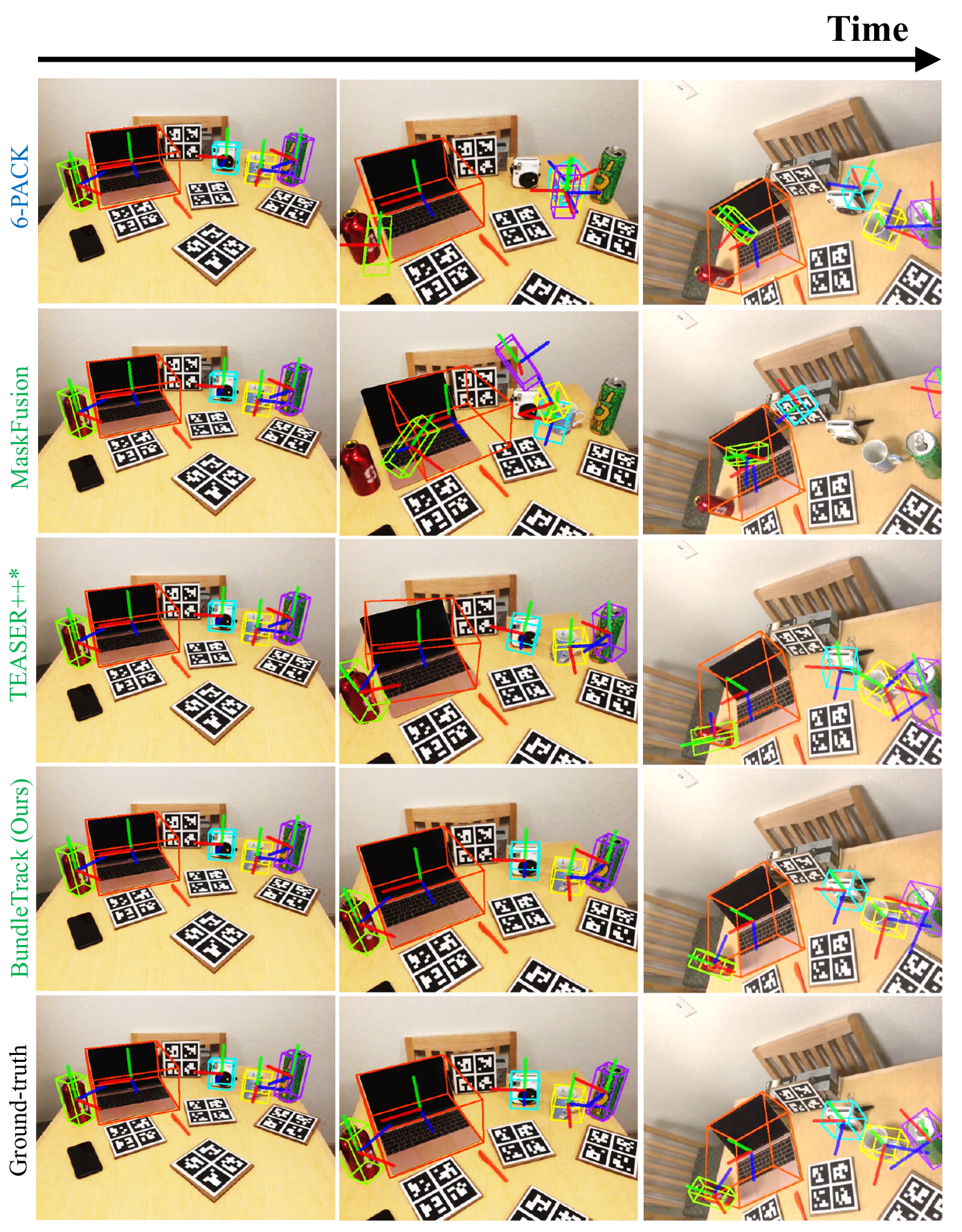}
  \vspace{-0.25in}\caption{Example qualitative results of \textit{BundleTrack} and representative comparison points on \textit{NOCS Dataset}. In all methods, each object is tracked individually and depicted in the same image for visualization. Methods' names are colored in \textcolor{blue}{blue} and \textcolor{green}{green} to denote assumption on \textit{category-level 3D model} and \textit{no model} respectively. For more qualitative results, please refer to the supplementary video.}
  \label{fig:nocs_qual}
  \vspace{-0.3in}
\end{figure}

\begin{figure}[h]
  \centering
  \vspace{+0.05in}
  \definecolor{red}{RGB}{255, 0, 0}
  \definecolor{blue}{RGB}{0,112,192}
  \definecolor{green}{RGB}{0, 176, 80}
  \includegraphics[width=0.48\textwidth]{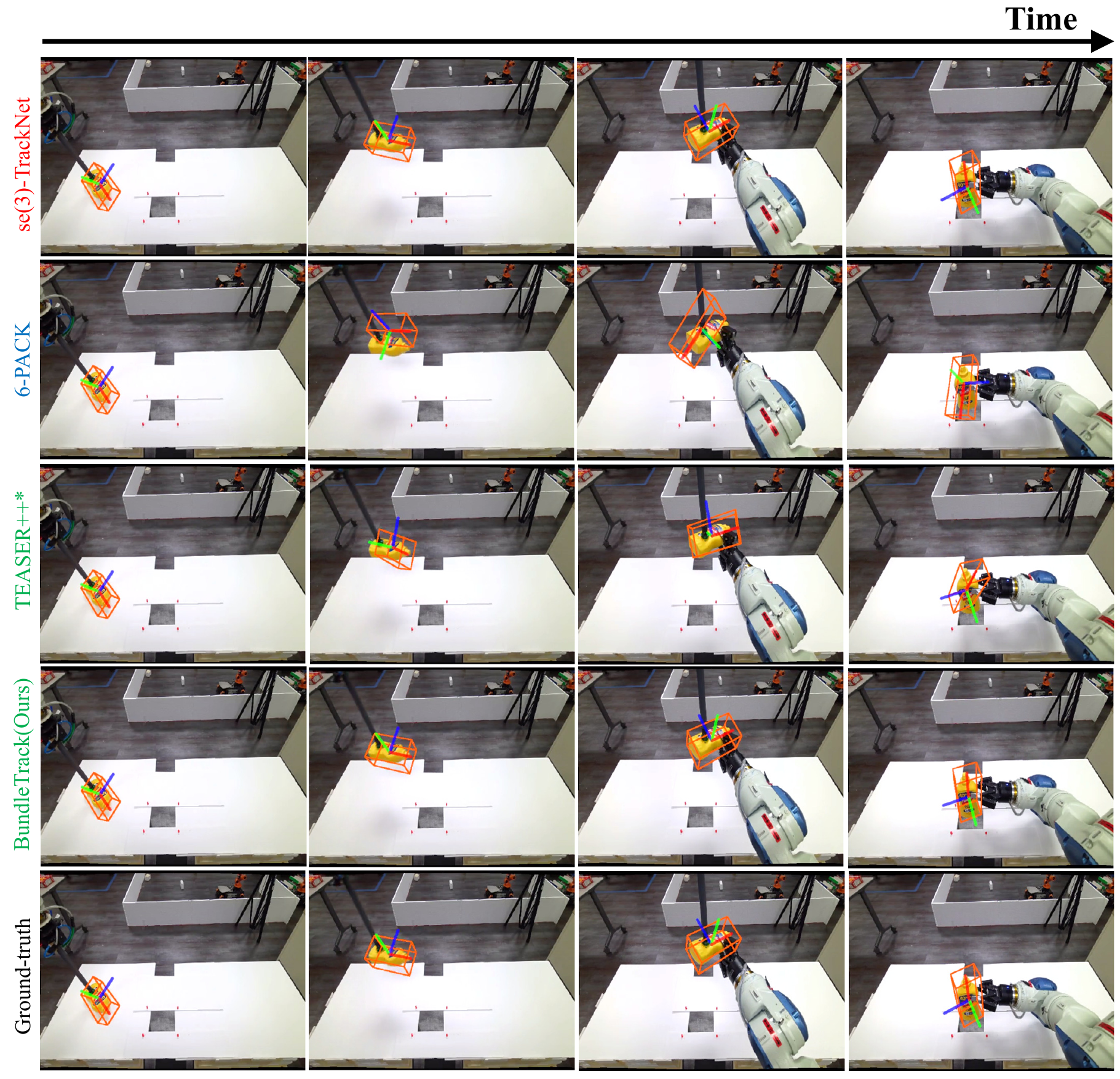}
  \vspace{-0.25in}\caption{Example qualitative results of \textit{BundleTrack} and representative comparison points on \textit{YCBInEOAT Dataset}. Methods' names are colored in \textcolor{red}{red}, \textcolor{blue}{blue} and \textcolor{green}{green} to denote assumption on \textit{instance 3D model}, \textit{category-level 3D model} and \textit{no model} respectively. For more qualitative results, please refer to the supplementary video.}
  \vspace{-0.3in}
  \label{fig:eoat_qual}
\end{figure}

Table \ref{tab:nocs} and Fig. \ref{fig:nocs_qual} present the quantitative  and qualitative results of state-of-art methods on the \textit{NOCS dataset} respectively. The comparison points include learning-based methods relying on a \textit{category-level prior}, such as \textit{NOCS} \cite{Wang_2019_CVPR}, \textit{KeypointNet} \cite{suwajanakorn2018discovery},  and \textit{6-PACK} with or without temporal prediction \cite{wang20196-pack}. These methods are offline trained on both real and synthetic training sets, which are rendered with 3D object models extracted from the same categories of \textit{ShapeNetCore} \cite{chang2015shapenet}.
In contrast, \textit{ICP} \cite{zhou2018open3d}, \textit{MaskFusion} \cite{runz2018maskfusion}, \textit{TEASER++\textsuperscript{*}} \cite{Yang20troteaser} and the proposed \textit{BundleTrack} have no access to any training data based on 3D models. 

The evaluation protocol is the same as in prior work  \cite{wang20196-pack}. A perturbed ground-truth object pose is used for initialization. The perturbation adds a uniformly sampled random translation within a 4cm range to evaluate robustness against a noisy initial pose \cite{wang20196-pack}. No re-initialization is allowed during tracking. To evaluate robustness against missing frames, the same uniformly sampled 450 frames out of 3200 in the testing videos are dropped \cite{wang20196-pack}. Four metrics are adopted: 1) \textbf{5\degree5cm}: percentage of estimates with orientation error < 5\degree and translation error < 5cm - the higher the better; 2) \textbf{IoU25} (Intersection over Union): percentage of cases where the overlapping prediction and ground-truth 3D bounding box volume is larger than 25\% of their union - the higher the better; 3) \textbf{R\textsubscript{err}}: mean orientation error in degrees - the lower the better; and 4) \textbf{T\textsubscript{err}}: mean translation error in centimeters - the lower the better. For  R\textsubscript{err} and T\textsubscript{err}, estimates with IoU$\leq$25 are not counted when computing averages\footnote{\href{https://github.com/j96w/6-PACK/blob/master/benchmark.py}{https://github.com/j96w/6-PACK/blob/master/benchmark.py}} \cite{wang20196-pack}. 
 
 
The results of comparison points other than \textit{MaskFusion} and \textit{TEASER++\textsuperscript{*}} come from the literature \cite{wang20196-pack}.  The open-sourced code\footnote{https://github.com/martinruenz/maskfusion} of \textit{MaskFusion} is used for evaluation, where the global SLAM module is disabled to avoid inferring object poses from the camera's estimated ego-motion. The dynamic object tracking module is kept to solely evaluate object pose tracking effectiveness. Its original segmentation module \textit{Mask-RCNN} \cite{he2017mask} is fine-tuned on the real training data provided in the \textit{NOCS dataset} for better performance while the synthetic data rendered using category-level 3D models are not used, as this method is also agnostic to any 3D models \cite{runz2018maskfusion}. In addition to \textit{ICP} reported in \cite{wang20196-pack},  another state-of-art 3D registration approach \cite{Yang20troteaser} is included for comparison and denoted as \textit{TEASER++\textsuperscript{*}}, which is robust to outlier correspondences and agnostic to 3D models. It takes as input the segmented point cloud and feature correspondences that are computed using the same modules proposed in \textit{BundleTrack}. 
For \textit{BundleTrack}, an initial mask $M_0$ is required as input to the framework and is provided via the aforementioned \textit{Mask-RCNN}. During execution, \textit{BundleTrack} does not require external mask input nor any form of re-initialization. As exhibited in Table \ref{tab:nocs}, \textit{BundleTrack} significantly outperforms the comparison points under all metrics and over all object categories, despite not accessing \textit{instance or category-level 3D models}.

\begin{table*}[htb!]
\centering
\vspace{+0.2cm}
\resizebox{0.9\textwidth}{!}{

\begin{tabular}{c|c|rr|rr|rr|rr|rr||rr}
\hline
\multirow{2}[4]{*}{\textbf{Assumption}} &
  \multirow{2}[4]{*}{\textbf{Methods}} &
  \multicolumn{2}{c|}{003\_cracker\_box} &
  \multicolumn{2}{c|}{021\_bleach\_cleanser} &
  \multicolumn{2}{c|}{004\_sugar\_box} &
  \multicolumn{2}{c|}{005\_tomato\_soup\_can} &
  \multicolumn{2}{c||}{006\_mustard\_bottle} &
  \multicolumn{2}{c}{ALL}
  \bigstrut\\
\cline{3-14} &
   &
  ADD &
  ADD-S &
  ADD &
  ADD-S &
  ADD &
  ADD-S &
  ADD &
  ADD-S &
  ADD &
  ADD-S &
  ADD &
  ADD-S
  \bigstrut\\
\hline
\multirow{3}[2]{*}{\textbf{Instance 3D Model}} &
  RGF \cite{issac2016depth} &
  34.78 &
  55.44 &
  29.40 &
  45.03 &
  15.82 &
  16.87 &
  15.13 &
  26.44 &
  56.49 &
  60.17 &
  29.98 &
  39.90
  \bigstrut[t]\\
 &
  dbot PF \cite{Wthrich2013ProbabilisticOT} &
  79.00 &
  88.13 &
  61.47 &
  68.96 &
  86.78 &
  92.75 &
  63.71 &
  93.17 &
  91.31 &
  95.31 &
  78.28 &
  89.18
  \\
 &
  se(3)-TrackNet \cite{wense3tracknet} &
  \textbf{90.76} &
  \textbf{94.06} &
  \textbf{89.58} &
  \textbf{94.44} &
  \textbf{92.43} &
  \textbf{94.80} &
  \textbf{93.40} &
  \textbf{96.95} &
  \textbf{97.00} &
  \textbf{97.92} &
  \textbf{92.66} &
  \textbf{95.53}
  \bigstrut[b]\\
\hline
\hline
\textbf{Category-Level 3D Model} &
  6-PACK \cite{wang20196-pack} &
  - &
  - &
  4.18 &
  18.00 &
  - &
  - &
  12.82 &
  60.32 &
  34.49 &
  80.76 &
  - &
  -
  \bigstrut\\
\hline
\hline
\multirow{3}[2]{*}{\textbf{No Model}} &
  MaskFusion \cite{runz2018maskfusion} &
  79.74 &
  88.28 &
  29.83 &
  43.31 &
  36.18 &
  45.62 &
  5.65 &
  6.45 &
  11.55 &
  13.11 &
  35.07 &
  41.88
  \bigstrut[t]\\
 &
  TEASER++\textsuperscript{*} \cite{Yang20troteaser} &
  63.24 &
  81.35 &
  61.83 &
  82.45 &
  51.91 &
  81.42 &
  41.36 &
  71.61 &
  71.92 &
  88.53 &
  57.91 &
  81.17
  \\
 &
  BundleTrack (Ours) &
  \cellcolor[rgb]{ .851,  .851,  .851}\textbf{85.07} &
  \cellcolor[rgb]{ .851,  .851,  .851}\textbf{89.41} &
  \cellcolor[rgb]{ .851,  .851,  .851}\textbf{89.34} &
  \cellcolor[rgb]{ .851,  .851,  .851}\textbf{94.72} &
  \cellcolor[rgb]{ .851,  .851,  .851}\textbf{85.56} &
  \cellcolor[rgb]{ .851,  .851,  .851}\textbf{90.22} &
  \cellcolor[rgb]{ .851,  .851,  .851}\textbf{86.00} &
  \cellcolor[rgb]{ .851,  .851,  .851}\textbf{95.13} &
  \cellcolor[rgb]{ .851,  .851,  .851}\textbf{92.26} &
  \cellcolor[rgb]{ .851,  .851,  .851}\textbf{95.35} &
  \cellcolor[rgb]{ .851,  .851,  .851}\textbf{87.34} &
  \cellcolor[rgb]{ .851,  .851,  .851}\textbf{92.53}
  \bigstrut[b]\\
\hline
\end{tabular}%

}
\vspace{-0.05in}
\caption{Results of AUC measured by ADD and ADD-S metrics on \textit{YCBInEOAT Dataset} \cite{wense3tracknet}. Under each type of 3D model assumption, best results are in bold. TEASER++\textsuperscript{*} denotes TEASER++ \cite{Yang20troteaser} operating over the same segmented point cloud and feature correspondences as in the proposed \textit{BundleTrack}.}
\vspace{-0.25in}
\label{tab:eoat}
\end{table*}


\subsection{Results on YCBInEOAT Dataset}
\label{sec:eval_eoat}

Evaluation exclusively on static objects captured by a moving camera cannot completely reflect the properties of a 6D pose tracking method \cite{wense3tracknet}. For this reason, the \textit{YCBInEOAT dataset} is chosen to evaluate tracking in scenarios where objects are moving in front of the camera.  The same evaluation protocol is followed as in prior work \cite{wense3tracknet}. Results are computed from accuracy-threshold AUC (Area Under Curve) measured by $ADD = \frac{1}{m}\sum_{x \in M} ||Rx+T-(\hat{R}x+\hat{T})||$, which performs exact model matching, and $ADD$-$S = \frac{1}{m}\sum_{x_1 \in M} \min_{x_2 \in M}||Rx_1+T-(\hat{R}x_2+\hat{T})||$ \cite{xiang2017posecnn} designed for evaluating symmetric objects. Similar to prior work \cite{wense3tracknet}, the ground-truth object's pose in the camera's frame is provided as initialization. No re-initialization is allowed during the tracking process.

\begin{figure}[ht]
  \centering
  \includegraphics[width=0.48\textwidth]{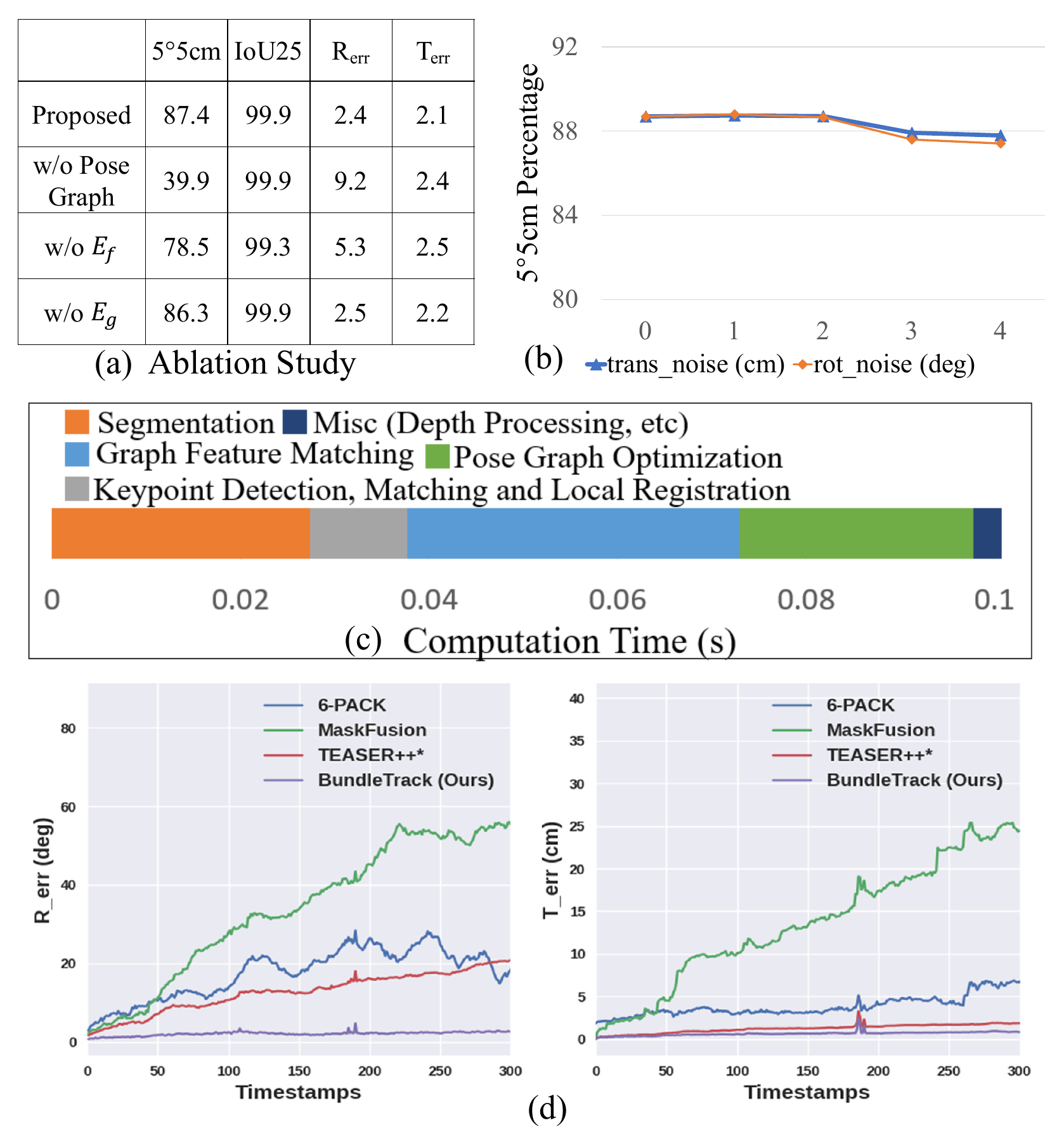}
  \vspace{-0.25in}\caption{Experimental analysis performed on \textit{NOCS dataset} as described in Sec \ref{sec:sys_analysis}. (a) Ablation study investigating effectiveness of pose graph optimization and each energy term. (b) Sensitivity of \textit{BundleTrack} to inaccurate initial pose by deliberately introducing different translation and rotation noise levels. (c) Average running time decomposition of different modules. (d) Rotation and translation error w.r.t. timestamps compared against representative related works \cite{wang20196-pack,runz2018maskfusion,Yang20troteaser} for tracking drift study.}
  \label{fig:analysis}
  \vspace{-0.3in}
\end{figure}

Quantitative and qualitative results are shown in Table \ref{tab:eoat} and Fig. \ref{fig:eoat_qual} respectively. Comparison points include state-of-art 6D pose tracking methods that use object CAD models, such as \textit{RGF} \cite{issac2016depth}, \textit{dbot PF} \cite{Wthrich2013ProbabilisticOT} and \textit{$se(3)$-TrackNet} \cite{wense3tracknet}. \textit{6-PACK} \cite{wang20196-pack} is a state-of-art 6D pose tracking approach relying on \textit{category-level 3D models}. Its evaluation on objects ``021\_bleach\_cleanser'', ``006\_mustard\_bottle'' and ``005\_tomato\_soup\_can'' are performed by using the officially released\footnote{https://github.com/j96w/6-PACK} networks trained on ``bottle'' and ``can'' category respectively . For the rest of the objects ``003\_cracker\_box'' and ``004\_sugar\_box'', no suitable corresponding category can be found in existing 3D model database \cite{chang2015shapenet} and thus \textit{6-PACK} is not able to be retrained and evaluated on them. For \textit{6-PACK}, 3D bounding box of the object model, computed from forward kinematics, is provided in every frame to crop ROI from point cloud, since it is more reliable than its default module of extrapolating the 3D bounding box by estimated motion.
For \textit{MaskFusion} \cite{runz2018maskfusion} and \textit{BundleTrack}, the initial object mask is obtained by table fitting and removal, followed by Euclidean Clustering implemented in PCL \cite{rusu20113d}. The original \textit{MaskFusion}'s segmentation module \textit{Mask-RCNN} cannot be retrained on this benchmark due to the lack of training set. Therefore, during tracking, the target object mask is computed by segmenting out the region of robot arm and end-effector from forward kinematics. For instances of irregular shapes or colors (``021\_bleach\_cleanser'', ``006\_mustard\_bottle'')  within the ``bottle'' category that \textit{6-PACK} has been trained on, it struggles to get satisfactory result. Nevertheless, \textit{BundleTrack} consistently demonstrates high quality tracking without any retraining or fine-tuning. This establishes generalizability of \textit{BundleTrack} to novel object instances regardless of their out-of-distribution properties within the category.  \textit{BundleTrack} also achieves comparable or superior performance even when compared against methods relying on object instance CAD models \cite{issac2016depth,Wthrich2013ProbabilisticOT,wense3tracknet}.

\vspace{-0.0in}
\subsection{Analysis}\label{sec:sys_analysis}
\vspace{-0.05in}
\noindent \textbf{Ablations Study:} An ablation study investigates the effectiveness of the online global pose graph optimization and each energy term, presented in Fig. \ref{fig:analysis} (a).

\noindent \textbf{Sensitivity to Initial Pose:} As mentioned, random translation noise within 4cm range is added to the initial pose. This part further investigates robustness under different translation and rotation noise levels, shown in Fig. \ref{fig:analysis} (b).

\noindent \textbf{Computation Time:} The average running time of modules are given in Fig. \ref{fig:analysis} (c). The entire framework runs at 10Hz on average including video segmentation. The \textit{6-PACK} \cite{wang20196-pack}, \textit{TEASER++\textsuperscript{*}} \cite{Yang20troteaser} and \textit{MaskFusion} \cite{runz2018maskfusion} methods from related work run at 4Hz, 11Hz and 17Hz respectively on the same machine.

\noindent \textbf{Tracking Drift Analysis:} Fig. \ref{fig:analysis} (d) presents the rotation and translation error w.r.t. timestamps compared against representative related works \cite{runz2018maskfusion,wang20196-pack,Yang20troteaser}. Results are averaged across all videos on the \textit{NOCS Dataset}. 

\noindent \textbf{Generalization:} The neural networks' weights and hyper-parameters in \textit{BundleTrack} are fixed without any retraining or fine-tuning across all evaluations (Sec. \ref{sec:eval_nocs}, \ref{sec:eval_eoat}). When applied to novel instances, the framework does not require access to \textit{instance or category-level 3D models} for training or registration.

\begin{figure}[h]
  \centering
  \includegraphics[width=0.48\textwidth]{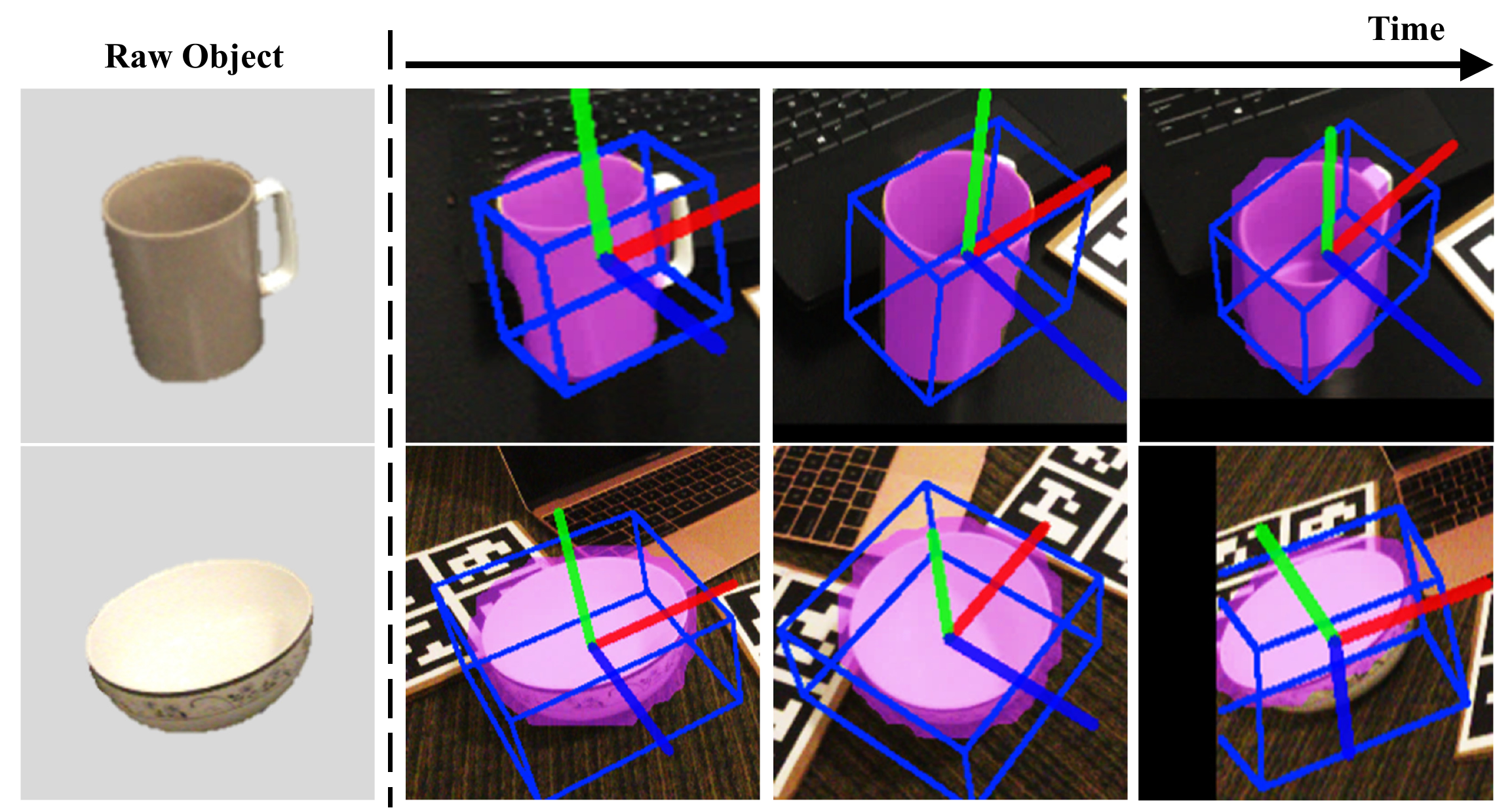}
  
  \vspace{-0.1in}\caption{Some of the most challenging cases for \textit{BundleTrack} on the \textit{NOCS Dataset}. \textbf{Top:} Severe self-occlusion prevents data association around the mug's handle, introducing challenges for solving the orientation around the green axis. Nevertheless, with better visibility in subsequent frames, \textit{BundleTrack} is able to recover from drifts and continue tracking, thanks to the memory-augmented pose graph optimization. \textbf{Bottom:} Near the end of video, noisy segmentation (purple mask) falsely ignores the side of the bowl, preventing relevant feature extraction and leads to slight translation offset. With future development of more advanced segmentation module, the overall tracking performance is expected to be boosted.}
  \label{fig:failure}
\end{figure}

\noindent \textbf{Failure Cases:}  While \textit{BundleTrack} is able to robustly keep tracking in all experiments without lost or re-initialization, intermediate imprecise estimates are observed, such as the cases illustrated in Fig. \ref{fig:failure}. 

\section{CONCLUSION}
\vspace{-0.05in}
This  work  presents \textit{BundleTrack}, a general framework for tracking the 6D pose of novel objects without any assumptions on \textit{instance or category-level 3D models}. Extensive experiments demonstrate that it is able to perform long-term accurate tracking under various challenging scenarios. It even achieves comparable performance to state-of-art methods that depend on the target object's CAD model. Future research includes the exploration of combining \textit{BundleTrack} with model-free grasping methods \cite{ten2017grasp, murali20206}, to perform robust pick-and-place \cite{morgan2021visiondriven,mitash2020task} or in-hand dexterous manipulation for a wide variety of novel objects.


\bibliographystyle{IEEEtran}
\bibliography{ref.bib}

\end{document}